# Knowledge Distillation-based Information Sharing for Online Process Monitoring in Decentralized Manufacturing System

Zhangyue Shi, Yuxuan Li, Chenang Liu[*]
School of Industrial Engineering and Management, Oklahoma State University, Stillwater, OK
*Corresponding author, chenang.liu@okstate.edu

**Abstract**: In advanced manufacturing, the incorporation of sensing technology provides an opportunity to achieve efficient *in-situ* process monitoring using machine learning methods. Meanwhile, the advances of information technologies also enable a connected and decentralized environment for manufacturing systems, making different manufacturing units in the system collaborate more closely. In a decentralized manufacturing system, the involved units may fabricate same or similar products and deploy their own machine learning model for online process monitoring. However, due to the possible inconsistency of task progress during the operation, it is also common that some units have more informative data while some have less informative data. Thus, the monitoring performance of machine learning model for each unit may highly vary. Therefore, it is extremely valuable to achieve efficient and secured knowledge sharing among the units in a decentralized manufacturing system for enhancement of poorly performed models. To realize this goal, this paper proposes a novel knowledge distillation-based information sharing (KD-IS) framework, which could distill informative knowledge from well performed models to improve the monitoring performance of poorly performed models. To validate the effectiveness of this method, a real-world case study is conducted in a connected fused filament fabrication (FFF)-based additive manufacturing (AM) platform. The experimental results show that the developed method is very efficient in improving model monitoring performance at poorly performed models, with solid protection on potential data privacy.

**Keywords**: Additive manufacturing, decentralized manufacturing system, *in-situ* process monitoring, knowledge distillation, machine learning.

# 1   Introduction

With the recent advancements of information technologies, more and more advanced manufacturing systems become cyber-enabled, which significantly facilitates information sharing among different units (e.g., sensors, machines, etc.) in an integrated system [1]. The incorporation of advanced online sensing technology in the manufacturing system provides valuable data to achieve efficient *in-situ* process monitoring. Signals from heterogeneous sensors could be utilized by applying appropriate data analytics methods [2, 3]. Accordingly, different types of *in-situ* monitoring methods have been developed to detect both internal and external quality issues [2, 4, 5]. Particularly, machine learning based methods gain more and more attention due to excellent monitoring performance.

Meanwhile, manufacturing systems also become connected and decentralized [6, 7]. In decentralized manufacturing system, it may occur that several manufacturers (i.e., units) fabricate same/similar type of product, collect corresponding data, and deploy machine learning model for process monitoring. On the one hand, some units may already fabricate a large number of samples while others may just start fabrication with less data available. On the other hand, due to equipment or operation difference, the data collected at different units may have different quality performance. Thus, the performance of machine learning models among different units may vary. As shown in *Figure 1*, different manufacturing units become connected and hence have opportunities to collaborate closely [8], which provides the potential to address the above issue by knowledge sharing. The unit that has more informative data in terms of either quality or quantity can be seen as a data-rich unit. Specifically, for the same type of product, the data-rich unit either has more collected data or has data with better quality. Thus, in general, these data-rich units are capable of training a machine learning model with superior performance. For other units who have less training data or have worse data quality, i.e., data-poor units, they may not have the same capability to train a satisfactory machine learning model. To improve the monitoring performance of data-poor units in the connected and decentralized environment, a natural idea is to seek help from data-rich units through knowledge sharing.

Therefore, the objective of this study is to enhance the *in-situ* process monitoring performance at data-poor units in the decentralized manufacturing system.

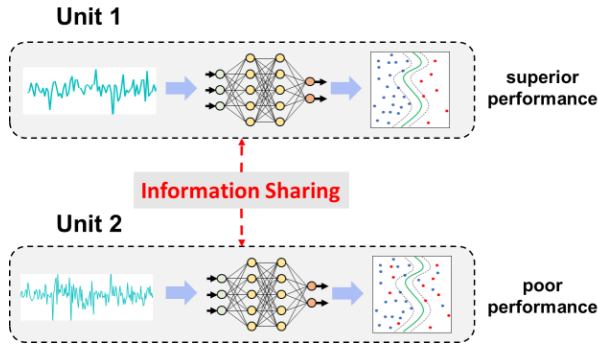

*Figure 1: A demonstration of decentralized connected manufacturing system.*

However, there are several technical challenges to realize this objective: (1) data collected from different units may follow different distributions, which makes knowledge sharing difficult. Although all units fabricate the same type of product, the data follow similar but not identical distribution due to the difference of working condition (machine, material, operation, etc.); and (2) data privacy needs to be well protected since some units may not want to directly share their valuable data with other units. To resolve these challenges, this study develops a knowledge distillation-based information sharing (KD-IS) framework to effectively share the informative knowledge learnt by the data-rich units with data-poor units.

As a powerful model compression framework, knowledge distillation is capable of extracting informative "dark" knowledge from complicated teacher model and then share it with simple student model [9]. Although knowledge distillation is originally designed to simplify machine learning model [9, 10], this study establishes an efficient knowledge sharing framework among manufacturing units by effectively leveraging knowledge distillation in the decentralized manufacturing system. In fact, it has the potential to share useful information even if both teacher and student models have the same architecture. Using the distilled knowledge, supervised monitoring performance in data-poor units could be enhanced without concern of data privacy. The rest of this paper is organized as follows. In Sec. 2, related work is briefly reviewed, and corresponding gaps are identified as well. The proposed research methodology is elaborated in Sec. 3. To validate the effectiveness of proposed method, Sec. 4 presents the experimental case study in fused filament fabrication (FFF) containing platform setup and results analysis. Eventually, conclusions are summarized and future works are discussed in Sec. 5.

## 2 Literature review

As introduced in Sec. 1, this study is motivated by enhancing the performance of individualized online process monitoring in decentralized manufacturing system. Therefore, in Sec. 2.1, relevant study regarding *in-situ* process monitoring is reviewed, and then followed by knowledge sharing methods to improve model performance in connected system (Sec. 2.2). The gaps in the current literature are identified accordingly.

### 2.1 Online process monitoring in advanced manufacturing

In the past decade, process monitoring has been thoroughly investigated [11-13]. Different sensors could be fused according to different manufacturing processes and corresponding signals could be utilized for online process monitoring. For example, vibration [2, 14, 15], acoustic emission [16], and temperature [17] signals are widely applied to detect process anomalies such as layer thickness change, clogging of the nozzle, etc. Also, image data such as optical camera [3, 18] and CT-image [19] could be utilized to detect porosity, part failure, etc. In addition, point cloud data containing rich process information is helpful to detect process shifts [20]. Video-based process monitoring also becomes popular recently, which is capable of detecting hot-spot in laser AM [21] and printing path alteration in FFF [5].

Speaking of detection methodologies, various analytical methods have been tailored to different processes. Conventional statistics methods contain control charts [2, 5], Bayesian model [17], spatio-temporal regression [21]. In addition, domain-aware methods such as the fractal analysis [19], and dimension reduction methods including principal component analysis [5], linear discriminant analysis [16] and manifold learning [22] are widely applied as well. Recently, machine learning-based methods also gain more and more attention in manufacturing, which consists of supervised learning and unsupervised learning methods. For supervised learning, boosting [2], random forest [14, 20], and support vector machine [23] are the common choices. Artificial neural networks such as multi-layer perceptron (MLP) [24], convolutional neural network (CNN) [15, 18, 25], recurrent neural network (RNN) [26], and long short-term memory network (LSTM) [2] are another popular tools to implement effective process monitoring. In terms of unsupervised learning, isolation forest [27], autoencoder [2], clustering [28], and one-class support vector machine [2] are widely used.

Currently, there are also multiple studies investigating how to enhance the process monitoring performance in single/independent manufacturing system. One way to do this is data fusion, which collects signals from different sensors and monitors process from different perspectives [29]. Another popular way is to continual learning, which is capable of progressively learning, fine-tuning, and transferring knowledge from long time spans while retaining previously learnt knowledge [30]. The recent studies have already demonstrated the potential of continual learning to predict part quality and detect anomaly in manufacturing systems [31].

However, for the above-mentioned methods, they focus on the process monitoring in single/independent manufacturing unit rather than connected and decentralized manufacturing system, which does not utilize useful information from other units that could improve the monitoring performance. Therefore, how to effectively utilize useful information to improve monitoring performance is discussed in Sec. 2.2.

**2.2 Knowledge sharing in decentralized manufacturing systems**

There are several ways to share useful information (i.e., knowledge) for performance improvement in the decentralized manufacturing system. Federated learning [32] and transfer learning [33] have the potential to achieve this goal. Federated learning could train unit-wise model for each unit in decentralized system without sharing data and update a global model by parameter sharing, which could effectively protect data privacy of each client. Currently, there are some pioneering work to utilize federated learning framework in the manufacturing to detect anomaly [34] and predict failure [35]. However, vanilla federated learning framework needs units to train their model almost synchronously since it needs to update the global model. In the decentralized manufacturing application, sometimes this requirement is hard to satisfy since not all units fabricate products simultaneously. Federated learning framework does not take temporal order of data collection into consideration. Although asynchronous federated learning framework has been developed to deal with the issue, the non-identical data distribution could also lead to increasing pattern bias learnt by federated learning server [34].

Transfer learning is another way to share the knowledge which consists of several categories such as fine-tuning and multitask learning [33]. Fine-tuning transfers a well-trained model from other domains to manufacturing by only fine-tuning parameters of few layers. In some works, pre-trained CNN [36] and autoencoder models [37] are transferred into manufacturing system for process monitoring and fault diagnosis. For fine-tuning transfer learning, a well pre-train model is needed, which require a large and representative dataset and is hard to satisfy. Multi-task learning is another type of transfer learning which aims to jointly learn multiple close-related tasks, which is capable of leveraging knowledge learnt in one task to other tasks in order to improve generalization performance [38]. Recently some multi-task learning based methods have been developed in manufacturing area for process variable prediction [39, 40]. Data privacy is one big issue that hinders the application of multi-task learning. In multi-task learning framework, data needs to be shared for model training. However, in some cases the client may not want to share their valuable data with other units.

Knowledge distillation (KD) is a model compression technique proposed by Hinton *et al.* in 2015 [9], which could effectively train a student model with simple architecture from a teacher model with complicated architecture. State-of-art performance of current algorithms has been achieved by complicated model with expensive computational time and excessive memory usage. Therefore, KD is required to generate a tiny model from the original cumbersome model [41]. As displayed in Figure 2, KD consists of teacher-student architecture following a distillation process. The cumbersome model obtained is seen as a teacher while the simple model is seen as a student. Teacher network has a good learning capacity and is able to pass a supervisory information (knowledge) to student network with lower learning capacity, which increases student's generalization ability [9]. In KD, knowledge can be understood as a mapping from the input vectors to output vectors. For classification problem, revised class probability output from teacher model is used as labels of the data and sent to the student network during the training of student [42].

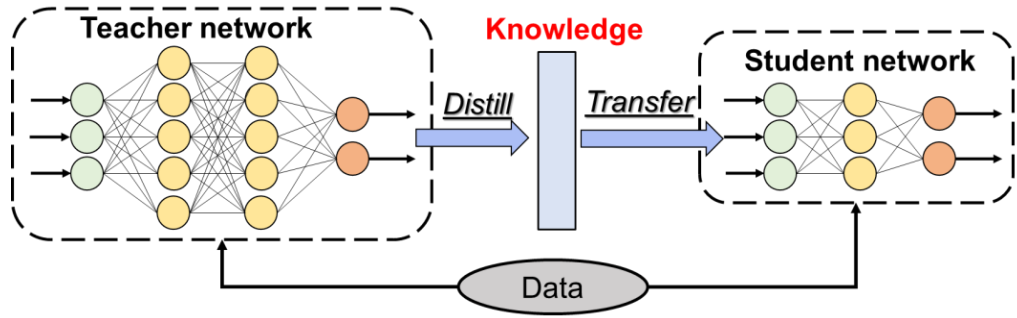

*Figure 2: Teacher-student framework for knowledge distillation.*

In general, logit-based or response-based knowledge denotes the response of the last output layer in teacher model. This type of knowledge is primarily proposed to mimic a teacher's final prediction [10]. This kind of knowledge is widely applied for various tasks such as visual recognition, speech recognition [42]. As part of KD, knowledge is transferred from a cumbersome teacher model to a simple student model. In addition to that, some follow-up works have been further extended based on the distilled model. A mutual learning framework [43] was proposed, which replaced fixed teacher-student architecture with a co-teaching architecture where a group of networks could guide each other. The Kullback Leibler (KL) divergence was also introduced to measure the similarities between the output probabilities of two networks. An early-stop distillation framework is proposed to make knowledge distillated from teacher network easier for student to learn by making teacher model less powerful [44]. A new loss function and a multi-exit architecture for ensemble KD was proposed by Phuong *et al.* to ensure the diversity of distribution [45]. Yang *et al.* kept using soft labels in the distillation process and added several constrains to optimize the model in successive generations [46]. A strategy to use a more noisy dataset is proposed to improve performance of student network by focusing on the data issue [47].

Currently, there are some studies investigating how to incorporate knowledge distillation framework in manufacturing to simplify the model [48-50] while utilizing knowledge distillation for information sharing to enhance monitoring performance needs to be explored [51]. Meanwhile, the teacher and student model are trained with the same dataset following same distribution [9], which needs to be adjusted in the decentralized manufacturing system because the data distribution at different units may not be identical. In addition, the architecture of teacher and student model does not have to be different in decentralized manufacturing system. Therefore, an information sharing framework is developed based on knowledge distillation in Sec. 3, which could narrow the gap of the existing solutions.

## 3 Research methodology

### 3.1 Problem formulation

In decentralized manufacturing system, each unit could collect their own data from side channels. For unit $i$, depending on specific sensor type, the $j$-th data are in form of $X_j^i$ where $X_j^i$ can either be a $k \times 1$ vector where $k$ denotes the dimension of side channel or a $k \times n$ matrix where $n$ is the window size. As displayed in Figure 3, the objective for each unit is to detect process anomaly during fabrication. Therefore, each unit collects its own data and then deploys own machine learning model $\Theta_i$ (neural network in this study) to detect process anomalies.

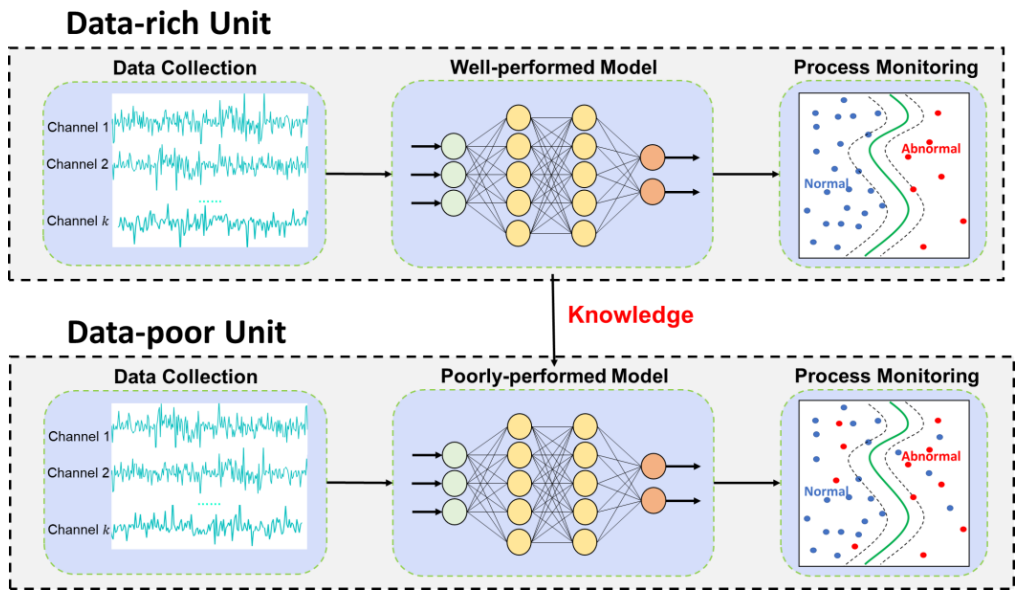

*Figure 3: Sharing process monitoring knowledge among different units in manufacturing system.*

In addition, it is worth noting that the quantity of sensor data and the quality of data could be different in the decentralized manufacturing systems. Some units could have better data (both normal and anomaly condition) in either quality or quantity. As a result, it is highly possible for these data-rich units to train a model that has superior monitoring performance. To preserve data privacy, it would also be helpful to share some extracted informative knowledge from well-performed model rather than directly share original data for performance enhancement of machine learning model at data-poor unit. As elaborated in Sec 3.2, knowledge distillation has the potential to achieve this goal.

### 3.2 Knowledge distillation

Knowledge distillation is a model compression technique proposed by Hinton *et al.* in 2015 [9], which could effectively train a smaller student model by learning from a larger teacher model. As shown in Figure 2, a general knowledge distillation framework consists of three major components: teacher-student network architecture, knowledge, as well as distillation algorithm. The main idea of knowledge distillation is that the student model mimics the teacher model in order to achieve a comparable or even a superior performance [10]. Teacher network usually has more complicated network architecture and learns from the data first. After the training of teacher network, the useful knowledge could be distilled from the teacher network. Hereafter, student network learns the knowledge distilled from teacher network to mimic teacher's behavior together with the training data. The core of knowledge distillation is how to simplify model while maintaining relatively good performance.

One promising and efficient way to distill knowledge is to mimic the logit output of last layer in the teacher model (i.e., teacher's prediction). Soft target, i.e., the probability that the input data $X_j$ belongs to each class, is proposed as the logit of teacher network. For a $M$-class classification problem, the soft target of teacher network $\Theta_t$ could be calculated by the softmax function as:

$$p_t^m(X_j, T) = \frac{\exp(z_t^m/T)}{\sum_{m=1}^{M} \exp(z_t^m/T)} \qquad (1)$$

Where $X_j$ denotes the $j$-th training sample, $M$ is the number of classes, $z_t^m$ is logit of last layer for the $m$-th class and $T$ denotes temperature factor which could control importance of each target. For example, a higher $T$ produces a softer probability distribution among different classes (assign probability with less difference to each class). Soft target contains informative dark knowledge of teacher network and could enhance performance of student network. Empirically, when the student model is very small compared to the teacher model, lower temperatures work better [9]. This is because a very small model might not be able to capture all the information when we raise the temperature. Therefore, soft target could be seen as knowledge extracted from teacher network and it could be transferred to student network by matching output of two networks. To measure the output similarity between teacher network $\Theta_t$ and student network $\Theta_s$, the Kullback Leibler (KL) divergence is often employed, which could be formulated in Eq. (2),

$$D_{KL}(\Theta_t||\Theta_s) = \sum_{j=1}^{N}\sum_{m=1}^{M} p_t^m(X_j,T)\log\frac{p_t^m(X_j,T)}{p_s^m(X_j,T)} \tag{2}$$

where $N$ is the total number of samples, $p_t^m(X_j,T)$ denotes the soft target of teacher network and $p_s^m(X_j,T)$ denotes the output of student network at temperature $T$. In addition, for classification task, the cross-entropy loss $L_c$ is used as another part of the overall objective function to measure the error between student network's prediction and the ground truth labels. The cross entropy of student network is shown in Eq. (3),

$$L_{C_s} = -\sum_{j=1}^{N}\sum_{m=1}^{M} I(y_j,m)\log(p_s^m(X_j,T=1)) \tag{3}$$

where $L_{C_s}$ denotes cross-entropy loss in student network and $I$ is an indicator function formulated as

$$I = \begin{cases} 1, & if \quad y_i = m \\ 0, & if \quad y_i \neq m \end{cases} \tag{4}$$

Therefore, the overall loss function for the student network $\Theta_s$ can be formulated as,

$$L_{\Theta_s} = \alpha D_{KL}(\Theta_t||\Theta_s) + (1-\alpha)L_{C_s} \tag{5}$$

where $\alpha$ is a hyperparameter that adapts the importance of these two loss functions. According to the experiment result [9], $\alpha$ should be much larger than $(1-\alpha)$ in order to achieve superior performance (usually 0.7). In this way, student network $\Theta_s$ learns to correctly predict the ground truth label of training data (i.e., via cross entropy loss) as well as mimic the soft target estimation of teacher network $\Theta_t$ (i.e., via KL divergence loss). The overall working principles is displayed in Figure 4. First, data are sent to teacher network to train teacher network $\Theta_t$ via backpropagation using cross entropy loss function $L_{C_t}$.

$$\Theta_t = \Theta_t + \lambda_t \frac{\partial L_{C_t}}{\partial \Theta_t} \tag{6}$$

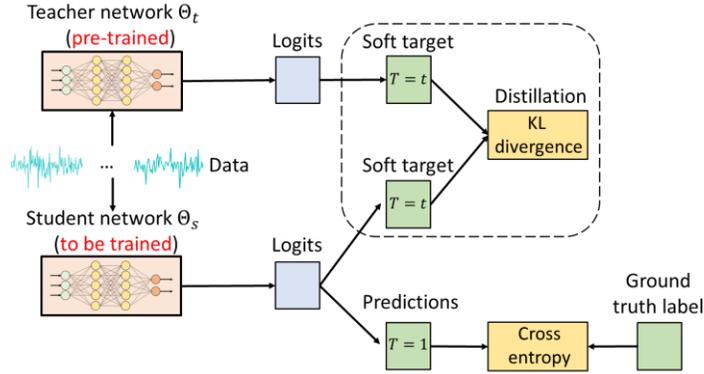

*Figure 4: The concrete architecture of knowledge distillation.*

When it comes to the training of student network, data are sent to the pre-trained teacher network in order to obtain a soft target at temperature $T = t$. Then student network is trained with the same dataset to compute both soft target at temperature $T = t$ and prediction at temperature $T = 1$. Afterwards, the objective loss $L_{\Theta_s}$ is calculated. Finally, student network parameters $\Theta_s$ are optimized via backpropagation and gradient descent using the overall loss function $L_{\Theta_s}$ containing both cross entropy and KL divergence.

$$\Theta_s = \Theta_s + \lambda_s \frac{\partial L_{\Theta_s}}{\partial \Theta_s} \tag{7}$$

The implementation details of knowledge distillation are presented in Algorithm 1.

**Algorithm 1: Knowledge Distillation**

*Input:* Training set $X$, label set $Y$, learning rate of teacher $\lambda_t$, learning rate of student $\lambda_s$
*Initialize:* Initialize teacher network $\Theta_t$ and student network's parameter $\Theta_s$; $k_t = 0$ $k_s = 0$;
*Repeat:*
    $k_t = k_t + 1$
    Choose mini-batch data $x$ from $X$
    Compute prediction of teacher $p_t(x, T = 1)$ by Eq. (1)
    Compute the gradient and update $\Theta_t$ using Eq. (6)
*Until:* Convergence
*Repeat:*
    $k_s = k_s + 1$
    Choose mini-batch data $x$ from $X$
    Compute soft target of teacher $p_t(x, T = t)$ by Eq. (1)
    Compute soft target and prediction of student $p_s(x, T = t)$, $p_s(x, T = 1)$ by Eq. (1)
    Compute the gradient and update $\Theta_s$ using Eq. (7)
*Until:* Convergence
*Output:* Student network $\Theta_s$

For the original knowledge distillation framework, the teacher network and student network are trained with the same data set, which follow the same data distribution. In practice, although different units fabricate identical type of product, the collected data may still follow different distributions due to the possible task inconsistency during the operation. Also, it is worth noting that different units may deploy neural networks with identical structure in the decentralized manufacturing system. Thus, necessary adaptions are critically needed to make knowledge distillation fit the decentralized manufacturing scenario, which is discussed in Sec. 3.3.

### 3.3 Knowledge sharing for online process monitoring in manufacturing system

To apply the knowledge sharing in decentralized manufacturing system, directly sharing the data among units is one option while it may lead to some privacy issues regarding the collected data. Knowledge distillation has the potential to enhance the model performance for data-poor units while preserving data privacy by sharing distilled knowledge rather than collected data. As introduced in Sec. 3.1, the amount of collected sensor data and the underlying distribution of data collected at each unit could be different in the decentralized manufacturing systems. In this section, a knowledge distillation-based information sharing (KD-IS) framework will be elaborated to enhance model performance of data-poor units in decentralized manufacturing.

*3.3.1 Knowledge distillation information sharing (KD-IS) in two-unit decentralized system*

Without loss of generality, the knowledge sharing approach is formulated with a cohort of two units in the decentralized manufacturing system, which could also be naturally extended to the multi-unit scenario [43]. Since the goal is to share useful knowledge among units rather than to simplify the model, in the decentralized manufacturing system, units are expected to have neural network with the same network architecture for process monitoring. Unit $i$ has $N_i$ samples $\boldsymbol{X}^i = (X_1^i, \dots, X_{N_i}^i)$ from $M$ different status. Correspondingly, the label set could be defined as $\boldsymbol{Y}^i = (y_1^i, \dots, y_{N_i}^i)$ where $y_j^i \in (1,2,\dots,M)$. And it is assumed data-rich unit has better data quality and/or quantity. When data-poor units start fabrication, it only has data with worse quality and/or quantity for both normal and anomaly status. Thus, student network $\Theta_s$ may have worse performance compared with teacher network $\Theta_t$. Since both units fabricate the same type of product, the knowledge learnt by data-rich unit could help to enhance the detection performance of the data-poor unit.

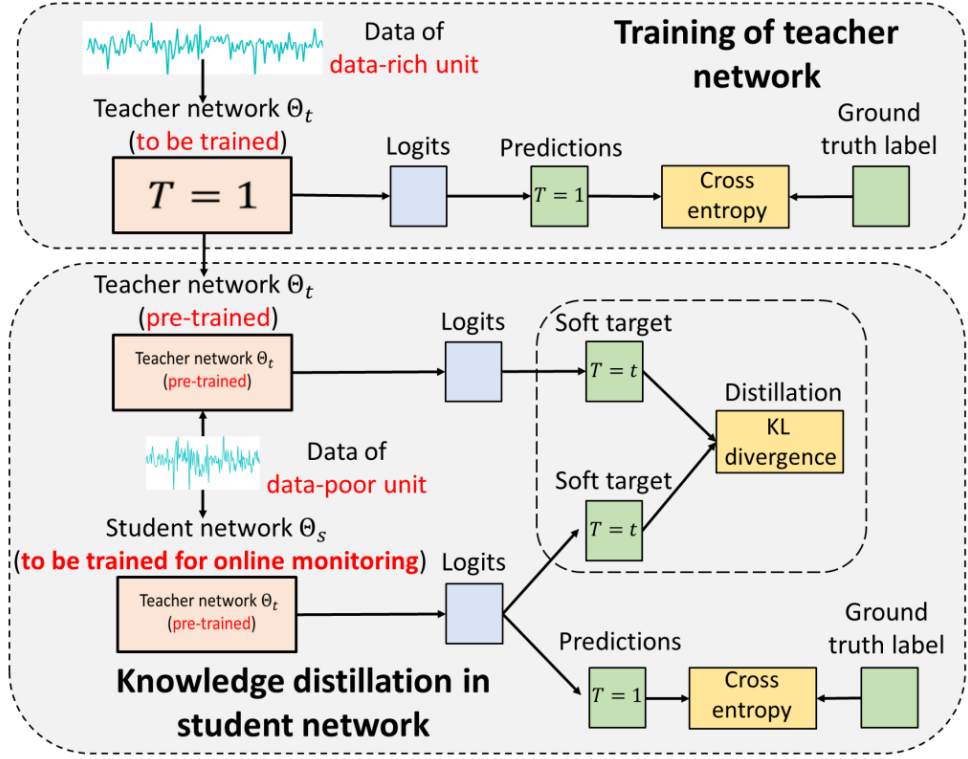

*Figure 5: The KD-IS framework in two-unit manufacturing system.*

Specifically, teacher network $\Theta_t$ should be determined first in the knowledge framework, which is based on the quality and quantity of the collected data. Then student network $\Theta_s$ needs to learn some knowledge from teacher $\Theta_t$. As displayed in Figure 5, teacher network $\Theta_t$ is trained first using data-rich unit's dataset $(X^t, Y^t)$ via backpropagation and minibatch as shown in Eq. (6). Then it comes to the training of student network $\Theta_s$. Different from vanilla knowledge distillation, student network $\Theta_s$ is trained with the data-poor unit's dataset $(X^s, Y^s)$ in decentralized manufacturing system. Similarly, learning strategy is embedded in mini-batches and $(x^s, y^s)$ are the data utilized in each mini-batch. The soft target of teacher network $p_t(x^s, T = t)$ is the knowledge that distilled from teacher network $\Theta_t$. Meanwhile, soft target $p_s(x^s, T = t)$ and prediction of student network $p_s(x^s, T = 1)$ are calculated in minibatch using Eq. (8).

$$p_s^m(x^s, T) = \frac{\exp(z_s^m/T)}{\sum_{m=1}^{M} \exp(z_s^m/T)} \qquad (8)$$

Then student's loss could be calculated using Eq. (5) and parameter of $\Theta_s$ could be updated via backpropagation as Eq. (7) displays. The optimization of $\Theta_s$ is conducted iteratively until convergence. According to the experiment results (see Sec 4.2), convergence can be observed with slower speed than trained with pure cross-entropy objective function. The detailed algorithm is summarized in Algorithm 2.

---

**Algorithm 2: KD-IS for Decentralized Manufacturing System**

*Input:* Unit network's training sets $(X^i, Y^i)$, learning rate $\lambda^i$
*Initialize:* Unit network's parameter $\Theta_i$;
*Step 1:* Determine teacher $\Theta_t$ and student $\Theta_s$ based on the data quality and quantity; set $k_t = 0$ $k_s = 0$
*Step 2:* Training teacher $\Theta_t$
    *Repeat:*
        $k_t = k_t + 1$
        Choose mini-batch data $(x^t, y^t)$ from $X^t$
        Compute prediction of teacher $p_t(x^t, T = 1)$ by Eq. (1)
        Compute the gradient and update $\Theta_t$ by Eq. (6)
    *Until:* Convergence
*Step 3:* Training student $\Theta_s$
    *Repeat:*
        $k_s = k_s + 1$
        Choose mini-batch data $(x^s, y^s)$ from $X^s$
        Compute soft target of teacher $p_t(x^s, T = t)$ by Eq. (1)
        Compute soft target and prediction of student $p_s(x^s, T = t)$, $p_s(x^s, T = 1)$ by Eq. (8)
        Compute the gradient and update $\Theta_t$ by Eq. (7)
    *Until:* Convergence
*Output:* Student network $\Theta_s$

---

It is worth noting that the role of teacher and student can switch at different scenarios. For example, unit 1 has better quality or more sufficient data for product 1 than unit 2. When training network to monitor process of product 1, knowledge learnt by unit 1 could be shared to unit 2. It is also possible that unit 2 has better or more sufficient data for product 2 than those of unit 1. In terms of process monitoring for product 2, unit 2 can be considered as the teacher and the detection performance of unit 1 could be enhanced by learning knowledge from unit 2. Compared with the conventional knowledge distillation where teacher and student model are trained with data following identical distribution, in the decentralized manufacturing system, the dataset in teacher (data-rich unit) and in student (data-poor unit) follows similar but not identical distribution as a result of the possible inconsistency in operation or machine. Essentially, both knowledge from teacher model and student's own dataset make contribution to training of student model.

*3.3.2 KD-IS for multi-unit decentralized manufacturing system*

In addition, KD-IS framework could be extended to multi-unit scenarios (see Figure 6). The data-rich unit 1 could train its own teacher model $\Theta_t$ first and then share its knowledge with data-poor units 2 and 3 during their model training. Therefore, in a connected decentralized system, for a specific task, any data-rich unit could share the knowledge learnt by its model to multiple data-poor units for the improvement of monitoring performance. And the role of teacher and student could switch depending on the specific tasks.

| **Algorithm 3: KD-IS for Multi-Unit Decentralized Manufacturing System** |
|---|
| ***Input:*** Unit network's training sets $(X^i, Y^i)$, learning rate $\lambda^i$ <br> ***Initialize:*** Unit network's parameter $\Theta_i$; <br> ***Step 1:*** Determine teacher $\Theta_t$ and student $\Theta_{s1}, \Theta_{s2}, ...,$ based on the data quantity; set $k_t = 0 \; k_s = 0$ <br> ***Step 2:*** Training teacher $\Theta_t$, the same as step 2 in Algorithm 2 <br> ***Step 3:*** Training student $\Theta_{s1}, \Theta_{s2}, ...,$ the same as step 3 in Algorithm 2 at each unit |

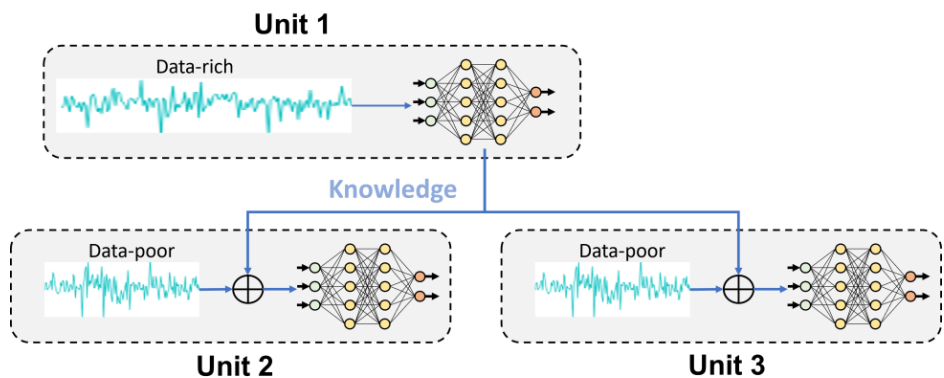

*Figure 6: The KD-IS framework in multi-unit manufacturing system.*

Incorporating KD-IS framework in the connected decentralized manufacturing system has several advantages. First, it could enhance the model performance of data-poor units. In addition, compared with directly sharing the data from data-rich unit to data-poor units, it could achieve competitive performance while taking less training time since it does not increase the amount of training samples for data-poor units.

Meanwhile, data may contain valuable process information so that sometimes data-rich units may not want to directly share data with other units [52]. In the KD-IS framework, only knowledge is shared with data-poor units rather than data, which could effectively protect the data privacy among different units. Also, the proposed KD-IS framework does not have conflicts with other existing knowledge sharing approaches such as federated learning and multi-task learning. Compared with federated learning, the proposed method focuses on the knowledge sharing from data-rich unit to data-poor unit while federated learning tries to improve all units in the system. In terms of multi-task learning, it could also be incorporated in the KD-IS framework. When some units within the decentralized system learn new tasks via multi-task learning, it could be considered as teacher model and corresponding knowledge could be distilled to enhance the performance of other units on the new task. In the next section, a case study in real-world application is conducted to validate the effectiveness of the proposed framework.

## 4  Case study

### 4.1  Experiment setup

As shown in Figure 7, two desktop FFF-based 3D printers (Anycubic Mega-S) are deployed for data collection. Each printer could be regarded as one unit. The vibration sensors are deployed to detect motion-related alteration during printing process. According to AM process knowledge, relative motion could be reflected by extruder [53]. Therefore, to detect AM process alterations, vibration sensors (i.e., MEMS accelerometers) are mounted to extruder of these two printers, respectively, which are capable of recording the real-time vibrations of the extruder in the three axes with approximately 1Hz sampling frequency. Raspberry Pi 4b microcontroller was used for data acquisition from all the side channels (i.e., vibration sensors in this study).

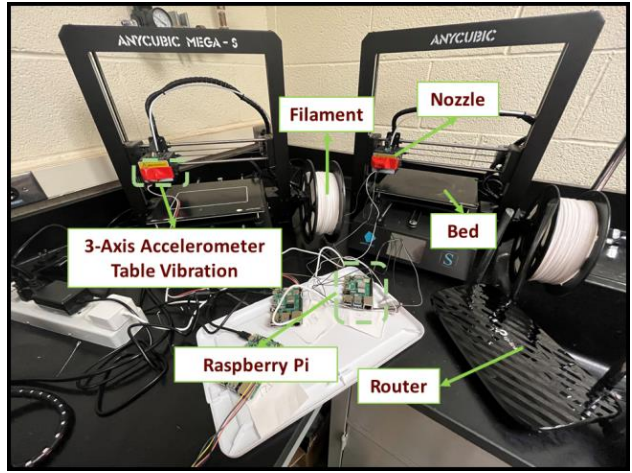

*Figure 7: Experimental platform: 2 Anycubic Mega S FFF printers.*

In this study, as shown in Figure 8 (a), a solid cube (with dimension $2\ cm \times 2\ cm \times 2 cm$) was used as the nominal design with the machine parameter setting summarized in TABLE 1. The printing material used in the experiment was polylactic acid (PLA) filament. To validate the effectiveness of the developed approach, a case based on potential cyber-physical attacks in AM was designed in this study, the detailed designed parameters is also shown in TABLE 1 as well. The design geometry could be attacked in "STL" or "G-code" stage. A small square-shaped void was maliciously inserted in the cube (see Figure 8 (b)), which is a simulated scenario where an internal feature (i.e., void) is inserted but cannot be detected by traditional quality inspection methods [2, 53]. However, this process anomaly could lead to compromised mechanical properties of the final product. Meanwhile, it is worth mentioning that the build time of nominal and defected products are comparable, so that the attack cannot be detected by simply tracking the printing time. To conduct the case study, 5 trials were performed in each printer and corresponding sensor data are collected lasting about 15 minutes for each trial, i.e., time to fabricate a test sample.

Since data contains temporal relationship, a natural idea is to analyze the stream data using a time window (see *Figure 9*). As defined in Eq. (9), for each unit $i$, the raw stream data could be reorganized as a sequence of overlapping time windows $X_j^i$ ($i$ denotes unit and $j$ denotes the time index of each window) with dimension $k \times n$,

$$X_j^i = \begin{pmatrix} x_{(j-1)\times(n-v)+1}^i \\ x_{(j-1)\times(n-v)+2}^i \\ \dots \\ x_{(j-1)\times(n-v)+n}^i \end{pmatrix}^T \tag{9}$$

where $n$ is the window size and $v$ denotes the number of overlapping data between two consecutive windows. In practice, the selection of window size $n$ and the number of overlapping data $v$ are two hyperparameters that can be empirically determined [54]. Each $X_j^i$ has a label representing current working status $y_j^i$.

*TABLE 1: The design parameters of nominal parts and altered parts.*

| Design Parameters | Nominal | Anomaly |
|---|---|---|
| Layer Design | Layer # 1~80: solid | Layer # 1~34: solid **Layer # 35~80: a square hole inside** |
| Printing Speed | 50 mm/s | 50 mm/s |
| Layer Thickness | 0.25 mm | 0.25 mm |
| Nozzle Temperature | 200 °C | 200 °C |
| Infill Rate | 20 % | 20 % |
| Bed Temperature | 60 °C | 60 °C |

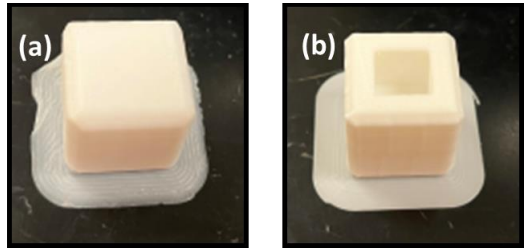

*Figure 8: Sample parts, (a) a nominal part, (b) an attacked part, a small square-shape void was maliciously inserted.*

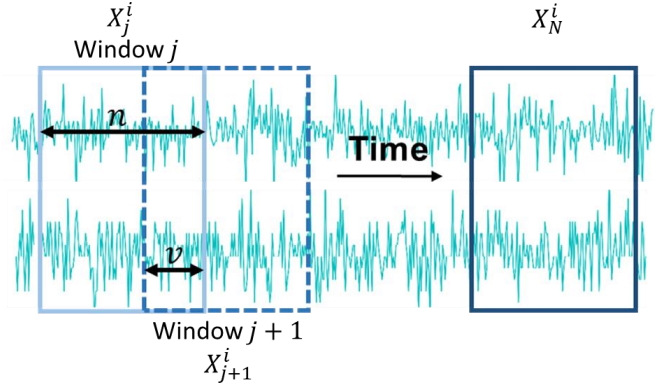

*Figure 9: Window-based data collection.*

**4.2 Results and discussions**

As described in Sec. 4.1, the side channel data collected from two different printers simulating potential process anomalies were applied to validate effectiveness of KD-IS framework. Specifically, two cases were conducted.

- **Case 1:** In the first case, data collected from printer 1 are used to train teacher network and data collected from printer 2 are used to train student network. The performance of student network is validated on the rest trials from printer 2.
- **Case2**: For the second case, the dataset for teacher and student network are switched. The teacher network is trained with data collected from printer 2 and student network is trained and validated on data collected from printer 1.

In terms of teacher network's training, more trials are used to train teacher network since data-rich unit has more available data. During each fold, to simulate data insufficiency situation in data-poor unit, student network is only trained using data of one trial and the rest four trials are used as testing set. A five-fold cross validation is applied on the student net.

*4.2.1 Parameter setup and benchmark methods*

In the experiment, a window-based sampling method is used to increase the number of samples. Preprocessing (i.e., standardization) is applied to the samples. The network selected in the study is 1D convolutional neural network as its ability to capture temporal relationship [25]. To guarantee fairness of comparison, teacher and student network have the same architecture. Teacher network is trained using Eq. (6) and student network is trained using Eq. (7). The parameter setup is listed in TABLE 2. To guarantee convergence, training epoch of student network is set in a relatively large number.

Furthermore, to demonstrate the effectiveness of the proposed KD-IS, two benchmarks are selected. The first one is the student network trained without knowledge distillation framework. In practice, this network is also trained with the same dataset as the proposed student network and has the same network architecture. The only difference is whether the student network is trained with knowledge distillation or not. The second benchmark is teacher network's performance on the test set, which aims to compare the performance of teacher network and student network with knowledge distillation. In this study, code was performed in Google Colab under Python 3.7.13 and PyTorch 1.12.0 + cu113.

*TABLE 2: The design parameters of two altered cases.*

| Step | Parameter | Value |
| --- | --- | --- |
| Window setup | Window size $n$ | 30 |
| | Window overlap $v$ | 28 |
| Knowledge Distillation setup | Weight parameter $\alpha$ | 0.7 |
| | Distilled temperature $T$ | 15 |
| CNN setup | Kernel size | 3 |
| | Stride | 2 |
| | Padding | 1 |
| | Hidden convolutional layer # of filter | (1, 16) |
| | Hidden convolutional layer # of filter | (16, 32) |
| | Hidden dense layer | (256, 120) |
| | Hidden dense layer | (120, 84) |
| | Output layer | (84, 2) |
| | Teacher net training epoch | 5 |
| | Student net training epoch | Case 1: 50<br>Case 2: 70 |

*4.2.2    The evaluation and comparison of monitoring results*

The classification performance is evaluated by four metrics, namely, accuracy, precision, recall and f-score. Precision and recall represent the level of type I and type II error, respectively. F-score is a metric that takes both precision and recall into consideration [55], which is formulated below,

$$\text{F} - \text{score} = 2 \times \frac{\text{Precision} \times \text{Recall}}{\text{Precision} + \text{Recall}} \tag{10}$$

The classification result of Case1 and Case 2 are presented in TABLE 3 and TABLE 4, respectively, which demonstrates that the proposed method could achieve best classification performance in both cases. Although teacher network is trained using data from data-rich unit, it could still achieve a good performance on the test set collected from data-poor unit. This is strong evidence showing that there are similarities in the data distribution of these two datasets and knowledge distilled from teacher network is helpful for training of student network in this scenario. Besides, student network trained with knowledge distillation has a 4%-5% improvement regarding accuracy and F-score in both cases compared with student network without knowledge distillation, which implies that the knowledge distillation framework could enhance the detection performance of network significantly.

As displayed in Figure 10, it could be observed that, after convergence, student network with knowledge distillation has better performance compared with student network without knowledge distillation. Also, it is worth noting the difference between knowledge distillation in our work and traditional knowledge distillation. In previous work [9], student and teacher are trained in the same dataset, where data follow identical distribution. However, in our study, the student and teacher network are trained with data following similar but not identical distribution, which demonstrates that KD-IS also has capability to enhance the model performance in decentralized manufacturing system even though there are some inconsistencies in data distribution.

TABLE 3: *The comparison of classification results – Case 1*

| Methods | Accuracy | Precision | Recall | F-score |
|---|---|---|---|---|
| Teacher network | 0.782 | 0.768 | 0.82 | 0.793 |
| Student network without KD | 0.764 | 0.811 | 0.708 | 0.749 |
| **Student network with KD (Proposed)** | **0.805** | **0.808** | **0.810** | **0.809** |

TABLE 4: *The comparison of classification results – Case 2*

| Methods | Accuracy | Precision | Recall | F-score |
|---|---|---|---|---|
| Teacher network | 0.778 | 0.810 | 0.738 | 0.772 |
| Student network without KD | 0.741 | 0.729 | 0.821 | 0.765 |
| **Student network with KD (Proposed)** | **0.782** | **0.748** | **0.875** | **0.807** |

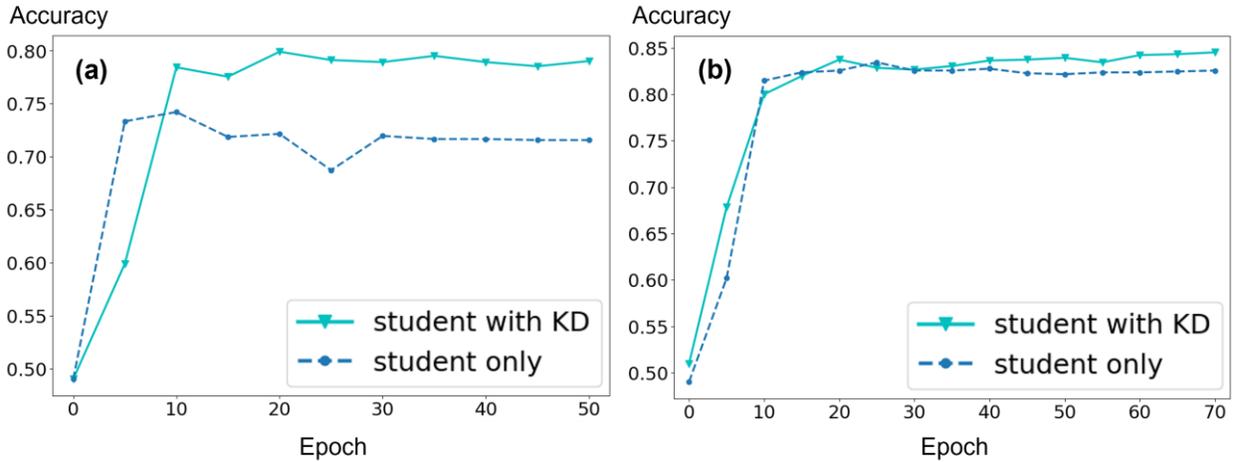

Figure 10: Student network's performance on the test set (a): One demonstration in case 1; (b): One demonstration in case 2.

*4.2.3 Model comparison with data-rich network*

Furthermore, we conducted an experiment that trains the same type of neural network with more data, i.e., data-rich network, which simulates the scenario that data-rich unit directly shares all data with data-poor units. Specifically, the data-rich network is trained using all available datasets except the test set. In case 1, the data-rich network is trained with all trials from printer 1 (all data in data-rich unit) and one trial from printer 2 (existing data in data-poor unit); then testified on the remaining four trials from printer 2. Similarly, in case 2, the training and test set are switched, where data-rich network is trained on all trials from printer 2 and one trial from printer 1; then validated on the remaining trials from printer 1. The setting is displayed in TABLE 5. For each case, a five-fold cross validation is performed as well. It is worth mentioning that these two networks also have the same architecture, and the only difference is the amount of training data. The data-rich network has much more training data (6 times) than the student network.

*TABLE 5: Parameter setting for investigation of network performance when trained with more data*

| Case | Name | Training set | Test set |
|---|---|---|---|
| Case 1 | Data-rich network | 5 trials in printer 1; 1 trial in printer 2 | 4 trials in machine 2 |
| | **Student network with KD (Proposed)** | 1 trial in printer 2 | |
| Case 2 | Data-rich network | 5 trials in printer 2; 1 trial in printer 1 | 4 trials in machine 1 |
| | **Student network with KD (Proposed)** | 1 trial in printer 1; | |

*TABLE 6: The comparison of classification results – Case 1*

| Methods | Accuracy | Precision | Recall | F-score |
|---|---|---|---|---|
| Data-rich network | 0.812 (0.032) | 0.838 (0.049) | 0.789 (0.075) | 0.810 (0.037) |
| **Student network with KD (Proposed)** | **0.805 (0.015)** | **0.808 (0.022)** | **0.810 (0.038)** | **0.809 (0.017)** |

*TABLE 7: The comparison of classification results – Case 2*

| Methods | Accuracy | Precision | Recall | F-score |
|---|---|---|---|---|
| Data-rich network | 0.818 (0.025) | 0.836 (0.022) | 0.804 (0.077) | 0.817 (0.036) |
| **Student network with KD (Proposed)** | **0.782 (0.044)** | **0.748 (0.053)** | **0.875 (0.080)** | **0.807 (0.035)** |

The average results of five-fold cross validation are shown in TABLE 6 and TABLE 7, respectively. The standard deviation is the value in the bracket. It could be observed that even though benchmark method utilizes much more data (6 times) for training, the accuracy and f-score of student network with knowledge distillation is still comparable. In addition, although benchmark uses much more data, the standard deviation of student network with knowledge distillation is much smaller in case 1 and is similar in case 2. Student network trained with knowledge distillation has good robustness even with less training data. Meanwhile, the computational cost is compared. In terms of training time, the proposed method spends 25 seconds in model training while benchmark needs 32 seconds to train the model since it has more training data. Student network saves about 25% of training time compared with the benchmark method, while maintaining a comparable model performance. It is also worth noting that the proposed method does not need to share data from data-rich client to data-poor unit, which could effectively avoid the potential data privacy issue as well. Therefore, it can be concluded that the proposed method is very promising to handle the process monitoring in decentralized manufacturing system.

## 5    Conclusions and future work

This paper develops a knowledge distillation-based information sharing (KD-IS) framework to enable the mutual assistance of machine learning-based online process monitoring among units in decentralized manufacturing systems. With the integration of knowledge distillation, the developed KD-IS is capable of distilling useful information from the data-rich unit (more knowledge available) to help data-poor units. Specifically, by learning the ground truth label together with soft target from teacher neural network of the data-rich unit, the student neural network at data-poor unit could have superior monitoring performance compared with student network without the assistance of knowledge distillation. In addition, the developed KD-IS also protects data privacy very well and computational cost is relatively low, as it does not require to share the raw data and completely retrain the model. A real-world case study using an FFF-based decentralized AM platform is designed to validate the effectiveness of the KD-IS. In summary, conclusions could be drawn as follows: (1) In a decentralized manufacturing system, data-poor unit could have a better monitoring performance by learning from the model of data-rich unit using proposed KD-IS framework; (2) the knowledge distillation mechanism works well in situation where teacher, student have the same network architecture and data follow similar but not identical distribution; and (3) data privacy issue could be eliminated by sharing distilled knowledge rather than directly sharing data.

This work is a preliminary study for enabling knowledge sharing among the manufacturing units in decentralized manufacturing systems. Several limitations still need to be addressed in the future work. On one hand, it is assumed that there is already a data-rich unit in current work. It is also possible that all units have limited training data. In the future, knowledge sharing without data-rich unit will be explored. On the other hand, the design of the test parts is relatively simple in this study. The robustness of the proposed method to complicated printing paths needs further investigation when geometry design becomes more complicated.